\documentclass[runningheads]{llncs}
\usepackage{graphicx}
\usepackage{comment}
\usepackage{amsmath,amssymb} 
\usepackage{color}

\usepackage[pagebackref=true,breaklinks=true,letterpaper=true,colorlinks,bookmarks=false]{hyperref}

\usepackage{mmstyle}
\usepackage{booktabs} 

\def \hfillx {\hspace*{-\textwidth} \hfill}

\usepackage{pifont}
\newcommand{\cmark}{\ding{51}}%
\begin{document}
\pagestyle{headings}
\mainmatter
\def\ECCVSubNumber{1277}  

\title{A Unified Framework for Shot Type Classification Based on Subject Centric Lens} 

\titlerunning{A Unified Framework for Shot Type Classification}
%
\author{Anyi Rao\inst{1} \and
Jiaze Wang\inst{1}\and
Linning Xu\inst{1}\and
Xuekun Jiang\inst{2} \and \\
Qingqiu Huang\inst{1} \and
Bolei Zhou\inst{1} \and
Dahua Lin\inst{1}}
\authorrunning{A. Rao et al.}
%
\institute{CUHK - SenseTime Joint Lab, The Chinese University of Hong Kong  \\	\and
Communication University of China\\
\email{\{anyirao, hq016, bzhou, dhlin\}@ie.cuhk.edu.hk, xkjiang@cuc.edu.cn \\ jzwang.cuhk@gmail.com, linningxu@link.cuhk.edu.cn} 
}
\maketitle


\begin{abstract}
\emph{Shots} are key narrative elements of various videos, \eg movies, TV series, and user-generated videos that are thriving over the Internet.
The types of shots greatly influence how the underlying ideas, emotions, and messages are expressed.
The technique to analyze shot types is important to the understanding of videos, which has seen increasing demand in real-world applications in this era. 
Classifying shot type is challenging due to the additional information required beyond the video content, such as the spatial composition of a frame and camera movement. To address these issues, we propose a learning framework  Subject Guidance Network (SGNet) for shot type recognition. SGNet separates the subject and background of a shot into two streams, serving as separate guidance maps for scale and movement type classification respectively. To facilitate shot type analysis and model evaluations, we build a large-scale dataset \textit{MovieShots}, which contains $46K$ shots from $7K$ movie trailers with annotations of their scale and movement types. Experiments show that our framework is able to recognize these two attributes of shot accurately, outperforming all the previous methods. 
\footnote[1]{The dataset and related codes are released \href{https://anyirao.com/projects/ShotType.html}{here} in compliance with regulations.}
\end{abstract}

\section{Introduction}
\label{sec:introduction}

 \begin{figure}[t!]
	\centering
  	\includegraphics[width=\linewidth]{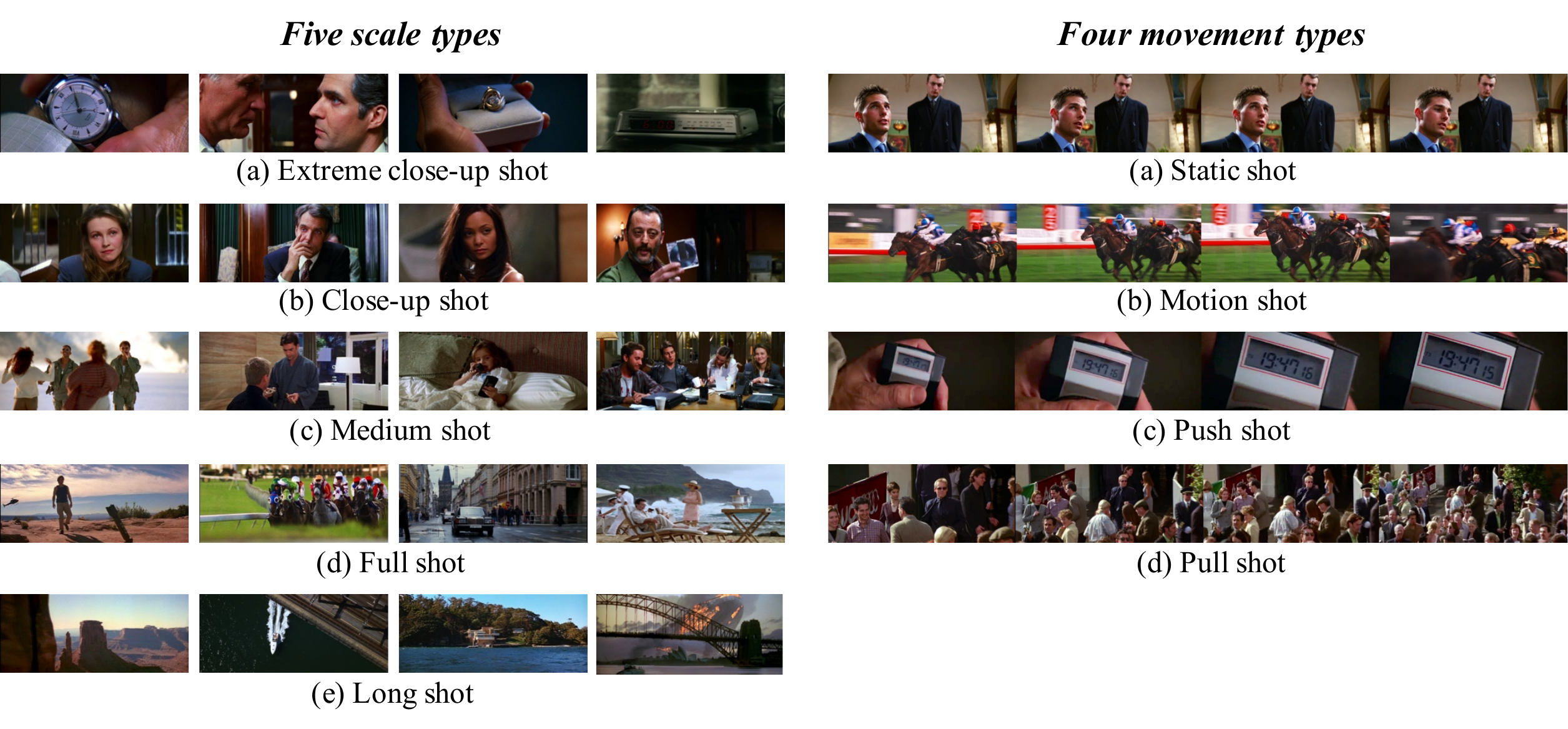}
	\caption{
		Demonstrations of five \emph{scale} types and four \emph{movement} types of video shots sampled from \emph{MovieShots} dataset. It is noticed that shot scales can reveal information of a story from different aspects. For example, \emph{long shots} usually indicate the location information, while \emph{close-up shots} are widely used for emphasizing the identities of the characters. \emph{Medium shots} and \emph{full shots} are good at depicting an event, while \emph{extreme close-up shots} are used for symbolic expressions or intensifying the story emotion. For movement types, we notice that \emph{static shots} are mainly used for narrative purposes and \emph{motion shots} try to track moving objects. \emph{Push shots} aim to emphasize the content information of the main subject while pull shots shrink the figure of the main subject and gradually reveal its surrounding environment
	}
	\label{fig:teaser}
\end{figure}

In 1900, film pioneer George Albert Smith firstly introduced shot type transitions into videos,  which revolutionized traditional narrative thought and this technique remains widely used in today's video editing~\cite{wiki}.
\emph{Shot}, a series of visual continuous frames, plays an important role in presenting the story. 
It can be recognized from multiple attributes, such as \emph{scale} and \emph{movement}.  
As illustrated in Fig.~\ref{fig:teaser}, five scale types and four movement types of shots are widely adopted in video editing, serving for different scenes and emotional expressions.

We are in the era of web 2.0 where user-generated videos proliferate, the techniques to analyze shot types have seen increasing demand in real-world applications: 1) With the capability of recognizing shot types, the videos shared online can be automatically classified or organized not only by their content, \eg object categories, but also by shot types. Thus the system will be able to respond to queries like “long shots over a city” etc. 2) By analyzing the sequence of shots in movies, we may provide a data-driven view of how professional movies are constructed. Such insight can help ordinary users to make videos that look more professional -- one can even build softwares to guide video production for amateurs.

Despite the potential value of shot type analysis, it is true that most previous works in computer vision primarily focus on objective content understanding. For example in video analysis, we focus on classifying and localizing the actions~\cite{wang2016temporal,caba2015activitynet,monfort2019moments}, while shot type analysis has been rarely investigated and lacks appropriate benchmark.
Existing datasets on shot type classification are either too small or not publicly available. However, we believe that the analysis of cinematic techniques (e.g. shot types) are also equally important. 

To facilitate researches along this direction, we construct a new dataset \emph{MovieShots}, which composes of over $46K$ shots collected from 
public movie trailers, with annotations on five scale types and four movement types. 
We select out {scale} and {movement} from many other shot attributes, as they are the two most common and distinguishable attributes that can uniquely characterize a shot in video, where the \emph{scale} type is decided by the amount of subjects  within the frame, and the \emph{movement} type is determined by the camera motion~\cite{giannetti1999understanding}. 

We further propose a novel subject centric framework, namely Subject Guidance Network (SGNet), to classify the scale and movement type of a shot. The key point here is to find out the dominant subject in a given shot, then we can decide its {scale} according to the portion it takes, and differentiate between the camera movement and subject movement to determine the {movement} type. SGNet successfully separates the subject and background in a shot and takes them to guide the full images to predict the labels for scale and movement type.

The contributions of this work are as follows:
1) We construct \emph{MovieShots}, a large-scale $46K$ shot dataset with professionally annotated scale and movement attributes for each shot.
2) SGNet is proposed to classify scale and movement type simultaneously based on the subject centric lens.
Our experiments show that this framework greatly improves the classification performance comparing to traditional methods and conventional deep networks TSN~\cite{wang2016temporal} and I3D~\cite{carreira2017quo}.


\section{Related Work}
\label{sec:related}
\noindent\textbf{Shot Type Classification Datasets.}
Traditional shot type classifications mainly focus on sports videos~\cite{ekin2003shot,duan2005unified,jiang2011tennis}. Sports video is a special kind of video that contains many clips such as video replays or comments, which is hard to transfer to general video scenarios.
Previous movie shot type researches~\cite{wang2009taxonomy,canini2013classifying} are limited on their evaluation benchmarks. They collect no more than twenty films with about one-thousand shots. There is no public available dataset to test the functionality of these methods. 
It is noticed that these datasets annotate either \emph{scale} or \emph{movement} attribute only, lacking a comprehensive description for a shot.
In order to solve these limitations, we collect a $10\times\sim100\times$ larger dataset, with $46K$ video shots annotated with both \emph{scale} and \emph{movement} attributes from more than $7K$ public movie trailers.\\

\noindent\textbf{Shot Type Classification Methods.}
Conventional methods for shot \emph{scale} classification use SVM with dominate color region~\cite{li2009soccer}, low-level texture features~\cite{xu2011using,bagheri2012new}, or optical flow~\cite{jiang2011tennis}. Decision tree method~\cite{2004icsp} sets up fixed rules to classify the scale type of a shot. Scene depth~\cite{benini2010estimating} is applied to infer the scale but is limited to the depth approximation accuracy and lacks generalization ability.
For \emph{movement} type classification, traditional approaches rely on the manual design of a motion descriptor. \eg~\cite{hasan2014camhid,prasertsakul2017video}~design \emph{motion vectors} CAMHID and 2DMH to capture the camera movement.
\cite{wang2009taxonomy}~leverage optical flow to find an alternative of the motion vectors.
However, all these methods heavily depend on hand-crafted features that are not applicable to general cases.
Our SGNet separates the subject from image and considers both the spatial and temporal configurations of a given shot, achieving much improved performance with better generalization ability.\\

\noindent\textbf{Video Analysis and Understanding in One Shot.}
Most previous single shot video understanding tasks~\cite{caba2015activitynet,monfort2019moments} are about action recognition~\cite{feichtenhofer2018slowfast,wang2016temporal,Wang_2018_CVPR} and temporal action localization~\cite{chao2018rethinking,xu2017r,zhao2017temporal}.
Video object detection~\cite{deng2019relation,guo2019progressive}, video object segmentation~\cite{zeng2019dmm,xu2019spatiotemporal}, video person recognition~\cite{yang2020spatial,xia2020online}, video-text retrieval~\cite{Xiong_2019_ICCV,yuan2020central}
and some low-level vision tasks, \eg~video inpainting~\cite{Xu_2019_CVPR} video super-resolution~\cite{zhang2019two,Li_2019_CVPR} are also applied in single shot videos.
However, research on video shot type is rarely explored, despite of its huge potential for video understanding.
We set up a benchmark with our \emph{MovieShots} {dataset} and conduct a detailed study on it.

\section{\emph{MovieShots} Dataset}
\label{sec:dataset}
To facilitate the shot type analysis in videos, we collect 
\emph{MovieShots},
a large-scale shot type annotation set that contains $46K$ shots from $7858$ movies. The details of this dataset are specified as follows. 

\begin{table}[t]
	\begin{minipage}{0.5\textwidth}
		\centering 
		\caption{Comparisons with other datasets}
		\resizebox{\textwidth}{!}
		{
			\begin{tabular}{lrrccl}
				\toprule
				& \#Shot & \#Video & Scale & Move.   \\ \midrule
				Lie 2014~\cite{lie}   
				&327   &  327     &          &   \cmark  & \\
				Unified 2005~\cite{duan2005unified}   
				&430   & 1        & \cmark         &     & \\
				Sports 2007~\cite{2007accv}   
				&1,364   &  8     &   \cmark &    & \\
				Soccer 2009~\cite{2009pcm}   
				&1,838   &  1     &   \cmark   &     & \\
				Cinema 2013~\cite{canini2013classifying}   
				&3,000  &  12     &   \cmark   &     & \\
				Context 2011~\cite{xu2011using}                  &3,206   & 4    & \cmark         &     & \\
				Taxon 2009~\cite{wang2009taxonomy}  
				& 5,054 & 7  &         & \cmark  &         \\\emph{MovieShots}                & 46,857       & 7,858  &  \cmark  &  \cmark           \\
				\bottomrule
			\end{tabular}
			\label{tab:datacomp}
		}
	\end{minipage}
	\begin{minipage}{0.5\textwidth}
		\centering
		\caption{Statistics of \emph{MovieShots}}
		\resizebox{\textwidth}{!}
		{
			\begin{tabular}{lrrrr}
				\toprule
				& Train 		&  Val 		& Test 	& Total\\ \midrule
				Number of Movies  		&  4,843  		&  1,062  		& 1,953    &7,858	\\
				Number of  Shots  		&  32,720  	& 4,610  	& 9,527 &46,857 \\ \midrule
				Avg. Dur. of Shot (s)  & 3.84    &  5.31    	& 3.78  &3.95\\
				\bottomrule
			\end{tabular}
			\label{tab:datastat}
		}
	\bigskip \bigskip \bigskip \medskip
	\end{minipage}
\end{table}


\subsection{Shot Categories}
Following previous definition on shot type~\cite{giannetti1999understanding,wang2009taxonomy,xu2011using,kowdle2012learning,savardi2018shot,huang2020movienet}, shot \emph{scale} is defined by the amount of subject figure that is included within the frame, while shot \emph{movement} is determined by the camera movement or the lens change.

Shot \emph{scale}  has five categories:
1) \emph{long shot} (LS) is taken from a long distance, sometimes as far as a quarter of a mile away;
2) \emph{full shot} (FS) barely includes the human body in full;
3) \emph{medium shot} (MS) contains a figure from the knees or waist up;
4) \emph{close-up shot} (CS) concentrates on a relatively small object, showing the face or the hand of a person; (5) \emph{extreme close-up shot} (ECS) shows even smaller parts such as the image of an eye or a mouth. 

Shot \emph{movement} has four categories:
1) in \emph{static shot}, the camera is fixed but the subject is flexible to move;
2) for \emph{motion shot}, the camera moves or rotates;
3) the camera zooms in for \emph{push shot}, and 4) zooms out for \emph{pull shot}.
While all the four movement types are widely used in movies, the use of \emph{push} and \emph{pull} shots only takes a very small portion. The usage of different shots usually depends on the movie genres and the preferences of the filmmakers. 

\subsection{Dataset Statistics}
\emph{MovieShots} consists of $46,857$ shots from $7,858$ movie trailers, covering a wide variety of movie genres to ensure the inclusion of all scale and movement types of shot.  
Table~\ref{tab:datacomp} compares \emph{MovieShots} with existing private shot type datasets, noting that none of them are publicly available.  \emph{MovieShots} is significantly larger than others in terms of the shot number and the video coverage, 
with a more comprehensive annotation covering both the \emph{scale} type and the \emph{movement} type for each shot.

\begin{figure}[!t]
	\centering
	\includegraphics[width=0.99\linewidth]{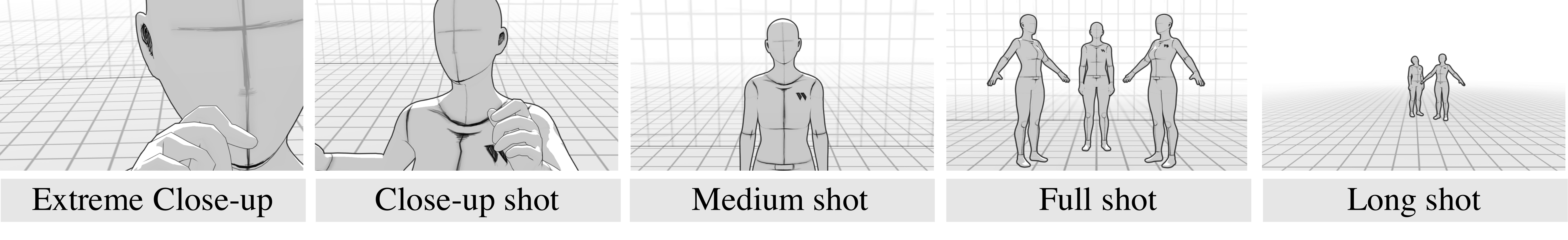}
	\caption{Prototypes of annotation corresponding to extreme close-up shot, close-up shot, medium shot, full shot and long shot}
	\label{fig:prototype}
\end{figure}

\subsection{Annotation Procedure}
Building a large-scale shot dataset is challenging in two aspects: appropriate data collections and accurate data annotations.
We firstly crawled more than $10K$ movie trailers online. Because movie trailers usually contain advertisements and big subtitles displaying the actor/director information, we firstly cleaned them with auto advertisement/big text detection and went through a second round of manual check. 
Noting that shot detection is a well solved problem and we used an off-the-shelf approach~\cite{sidiropoulos2011temporal,rao2020local} to cut shots in these trailers and filtered out failure cases with manual check.
All our annotators are cinematic professionals in film industries or cinematic arts majors, who provide high quality labels. We also set up annotation prototypes for these well defined criterion of shot types, as illustrated in Fig.~\ref{fig:prototype}.
Additionally, three rounds annotation procedures have been done to ensure the high consistency. We finally achieve $95\%$ annotation consistency, with those inconsistent shots being filtered out in our experiments.
\section{SGNet: Subject Guidance Network}
\begin{figure*}[!t]
	\begin{center}
		\includegraphics[width=\linewidth]{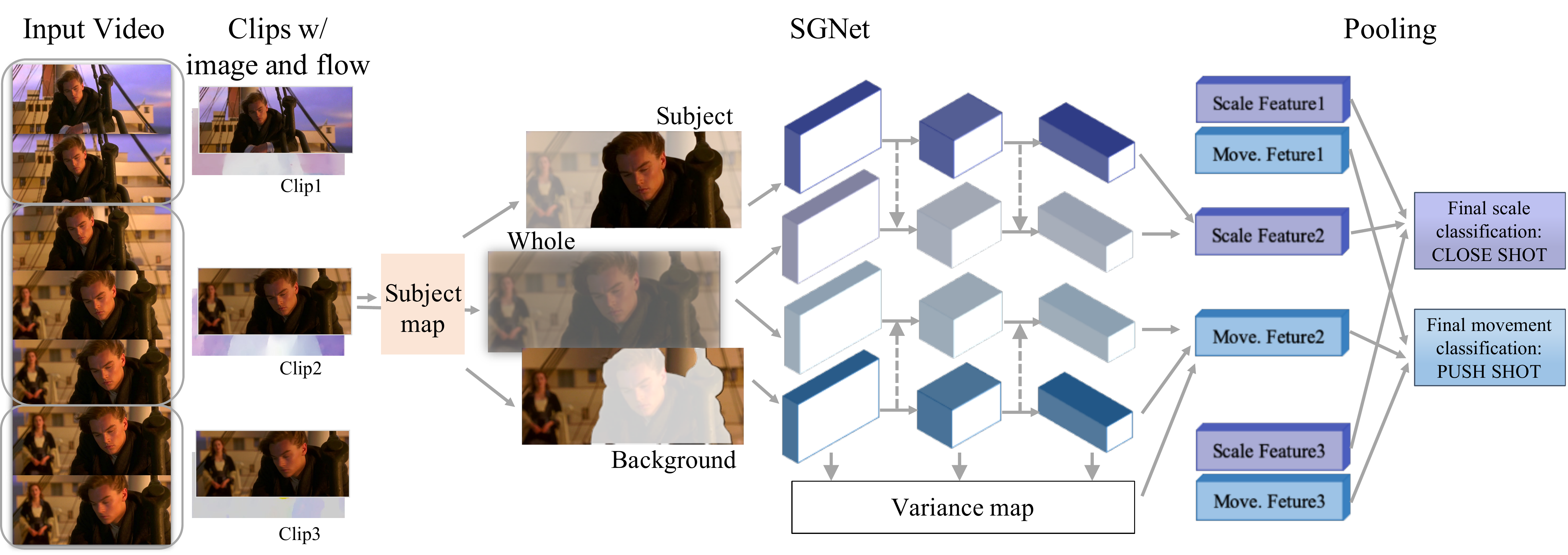}
	\end{center}
	\caption{
		Pipeline of Subject Guidance Network (SGNet). A subject map is generated from each clip's image. With this subject map, we take subject and background to guide scale and movement prediction respectively}
	\label{fig:pipeline}
\end{figure*}

In this section, we introduce our Subject Guidance Network (SGNet) for \emph{scale} and \emph{movement} type classification. The overall framework is shown in Fig.~\ref{fig:pipeline}.

We firstly divide a shot into $N$ clips to capture the contextual variations along the temporal dimension.
Each clip passes through a two-branch classification network and outputs a feature vector.
The feature vectors coming from the  $N$ clips are pooled together and pass through a fully-connected layer to get the final prediction.

It is noticed that, the separation of subject and background information is  critical for both two tasks. While the \emph{scale} type depends on the portion of the subject in the shot, the \emph{movement} type relies on the background motion rather than the subject motion, as the changes of the background information are closely related to the camera motion.
%
To reduce the burden of the whole pipeline,
a light-weight \emph{subject map generator} is designed to separate the subject and background\footnote[2]{Background image is equal to the whole image minus the subject part.} in both {image} and {flow} in an effective way.
The \emph{subject map} here is a saliency map sharing the same width and height as the original image with values in range of zero to one.
We use \emph{subject map} to guide the whole image to predict the \emph{scale}, and use the  \emph{background map} to guide the whole image to predict the \emph{movement} with the help of an obtained \emph{variance map}.

In the following two subsections, 
we will first introduce the subject map guidance for shot type classification in Section~\ref{sec:guide}, and elaborate on our subject map generation in Section~\ref{sec:atten}.

\subsection{Subject Map Guidance Classification}
\label{sec:guide}
As discussed before,
the separation of subject and background is crucial to shot type classification. 
In this section, we firstly explain how the obtained subject map guides scale and movement classification respectively.

For \emph{scale} type classification, the subject map $[1\times W \times H]$ element-wise multiplies with the whole image $[3\times W \times H]$ to get an subject image $[3\times W \times H]$.
Then we take the whole image and the obtained {subject} image as two input pathways. Each of them is sent into a ResNet50~\cite{he2016deep} network.
To apply subject guidance,  
we fues the subject feature map from different stages of feature representations  into the whole image pathway. 
Specifically, these fusion connections are applied right after
pool$_1$, res$_2$, res$_3$, and res$_4$ in the ResNet50~\cite{he2016deep} backbone. The fusion is conducted by lateral connections~\cite{christoph2016spatiotemporal}. 

For \emph{movement} type classification, the background guidance is applied in a similar way to the scale type prediction.
Additionally,  
a \emph{variance map} module is further introduced, inspired by the fact that the changes of appearance along time is a cue for movement classification. 
For example, in a \emph{static} shot, the appearances of background among different clips are almost the same, while the background appearances changes  significantly as time changes in \emph{motion} shots, \emph{push} shots and \emph{pull} shots.
We calculate one variance map $\mV_m \in \cR^{N\times N}$ for each shot among its different clips ($N = 8$ clips in our experiments) at different stages of the backbone ResNet50. Specifically, these stages include those after pool$_1$, res$_2$, res$_3$, and res$_4$, the same as those used in the previous fusions.
We apply inner product between the two normalized feature maps $\mF_{m,i},\mF_{m,j}\in \cR^{h_m\times w_m \times c_m}$ from two clips $i,j$ at stage $m$ to get $\mV_{m,(i,j)}$. The inner product here is equivalent to calculating the cosine similarity, which captures the similarities between different clips.
$\mV_{m}$ is achieved by concatenating all possible clips pair $\mV_{m,(i,j)}$. 
Finally, all $\mV = \{\mV_{m}\}$ among $M$ stages are concatenated along the channel-wise dimension and are fed into a two-layer FC for classification. The classification results using variance maps are fused with the image classification results for the final prediction.

\subsection{Subject Map Generation}
\label{sec:atten}
Now we elaborate on how we separate the subject from the background with our light-weight \emph{subject map generator}.

Conventional saliency/attention map methods employ hand-crafted visual features or heuristic priors~\cite{cheng2014global,zhu2014saliency}, which are incapable
of capturing high-level semantic knowledge, making the predicted map unsatisfactory.
Pre-trained state-of-the-art deep networks~\cite{li2016deepsaliency,deng2018r3net} are usually very large with more than $50\sim 100$ layers and are not easy to be taken as a submodule in the designed networks to fine tune, considering the high computational costs. 
From another perspective, to train a randomly initialized subject network from scratch with only shot type label is impractical, since the supervision signal is too weak and the network is unable to converge, considering the subject map is a pixel-wise prediction but the annotation is a video-level label.

To strike a balance between the performance and the computational efficiency, we resort to knowledge distillation (KD)~\cite{kd_hinton2015,xu2020knowledge}, considering that it is easy and flexible to learn, and achieves state-of-the-art performance on classification problems~\cite{cheng2017survey}.
A light-weight \emph{student generator} (a 6-layer CNN) learns from its \emph{teacher network} (a MSRA10K~\cite{HouPami19Dss} pre-trained $R^3$Net~\cite{deng2018r3net}).
Note that a naive KD using $\cL_2$ loss is usually suboptimal because it is difficult to learn the true data distribution from the teacher and may result in missing generation details. Therefore, an additional adversarial loss with the help of Generative Adversarial Networks (GAN)~\cite{KDGAN_NIPS2018,wang2018adversarial,shou2019dmc,belagiannis2018adversarial,goldblum2020adversarially} is adopted. 
In all, the student generator is trained by minimizing the following three-term loss,
$$
\cL = \alpha \mathcal{L}_{2}+\beta \mathcal{L}_{\mathrm{adv}} + \mathcal{L}_{\mathrm{cross}}.
$$
The first loss term $L_{2}$ is the least square error between the generated subject map and its corresponding pseudo subject map, which aims to mimic the output of teacher network. 
$L_{2}$ loss alone is not able to teach the student network to generate fine grained details since it does not consider the constraints from the whole data distribution.
The second term $\mathcal{L}_{\mathrm{adv}}$ is given by a learned discriminator, which is trained to compete with the student generator to learn the true data distribution. 
The discriminator takes the subject map from the teacher network as \emph{real} and the output from student generator as \emph{fake}.
%
Finally, we take cross-entropy loss $L_{cross}$ as our classification loss and be jointly trained with the whole pipeline to encourage right predictions.

\section{Experiments}
\subsection{Experiments Setup}

\paragraph{\bf Data.} All the experiments are conducted on \emph{MovieShots}.
The whole dataset is split into \textit{Train}, \textit{Val}, and \textit{Test} sets with a ratio 7:1:2, as shown in Table~\ref{tab:datastat}. 

\paragraph{\bf Implementation Details.}
We take cross-entropy loss for the classification result. The shot is evenly split into 3 clips in training and 25 clips in testing.
The fusing function from the subject/background map to whole image is implemented by concatenating the output from the two branches.
Image and flow are set up as two inputs and their classification score are fused to get the final results.
We train these models for $60$ epochs with mini-batch SGD, where the batch size is set to 128 and the momentum is set to 0.9. 
The initial learning rate is $0.001$ and the learning rate will be divided by $10$ at the $20th$ and $40th$ epoch.

\paragraph{\bf Evaluation Metrics.}
We take the commonly used Top-1 accuracy as the evaluation metric.
Specifically, in our experiment, we denote Acc$_{S}$ for scale classification performance and Acc$_{M}$ for movement classification performance.

\subsection{Overall Results}
We reproduce DCR~\cite{li2009soccer}, CAMHID~\cite{hasan2014camhid} and 2DMH~\cite{prasertsakul2017video} according to their papers.
DCR~\cite{li2009soccer} clusters dominant color sets and predicts shot type based on the ratio of different color sets.
CAMHID~\cite{hasan2014camhid}, 2DMH~\cite{prasertsakul2017video} are based on motion vectors. CAMHID~\cite{hasan2014camhid} takes SVD to get the dominant components. 2DMH~\cite{prasertsakul2017video} disentangles the magnitude and orientation of motion vectors. 
TSN~\cite{wang2016temporal} and I3D~\cite{carreira2017quo} are experimented using authors' code repositories. SGNet adopts ResNet50~\cite{he2016deep} as the backbone. All the network weights are initialized with pre-trained models from ImageNet~\cite{russakovsky2015imagenet} unless specially stated.\\

\begin{table}[!t]
	\begin{center}
		\caption{
			The overall results on shot scale and movement type classification
		}
			\resizebox{0.65\columnwidth}{!}{%
		\begin{tabular}{l|c|c}
			\toprule
			Models & Acc$_{S}$ ($\uparrow$)& Acc$_{M}$ ($\uparrow$) \\ \midrule
			DCR, Li \etal~\cite{li2009soccer}          & 51.53     & 33.20      \\
			CAMHID, Wang \etal~\cite{hasan2014camhid}       & 52.37     & 40.19      \\
			2DMH, Prasertsakul \etal~\cite{prasertsakul2017video}       & 52.35     & 40.34      \\
			\midrule
			I3D-ResNet50~\cite{carreira2017quo}     & 76.79  &  78.45          \\
			I3D-ResNet50-Kinetics~\cite{carreira2017quo}     & 77.11  &  83.25          \\
			TSN-ResNet50 (img)~\cite{wang2016temporal}      & 84.08     & 70.46  \\
			TSN-ResNet50-Kinetics (img)~\cite{wang2016temporal}      & 84.18     & 71.61  \\
			TSN-ResNet50 (img + flow)~\cite{wang2016temporal}      & 84.10     & 77.13    \\
			TSN-ResNet152 (img + flow)~\cite{wang2016temporal} & 84.95	&78.02\\
			\midrule	\midrule					
			SGNet (img) & 87.21     & 71.30      \\
			SGNet (img + flow)        &   87.50  &  80.65         \\
			SGNet w/ Var (img)  &   87.42  &  80.57    \\
			SGNet w/ Var (img + flow) &  \textbf{87.57}  & \textbf{81.86}    \\
			SGNet w/ Var-Kinetics (img + flow) &  \textbf{87.77}  &  \textbf{83.72}    \\
			\bottomrule
		\end{tabular}
			}
		\label{tab:result}
	\end{center}
\end{table}

\noindent\textbf{Overall Results Analysis.}
1) \textit{Traditional Methods.}
The overall results are shown in Table~\ref{tab:result}. The performances of DCR~\cite{li2009soccer}, CAMHID~\cite{hasan2014camhid} and  2DMH~\cite{prasertsakul2017video} are restricted by their poor representations. 

2) \textit{3D Networks.}
For movement classification, I3D-ResNet50 achieves better result than TSN-ResNet50 (img + flow) since it captures more temporal relationships. With Kinetics400~\cite{carreira2017quo} pre-trained, 
I3D-ResNet50 gets 4.8 boost on $Acc_M$.
But in scale classification, I3D-ResNet50 performs worse than TSN-ResNet50 (img). The reason might be that I3D-ResNet50 is not good at capturing the spatial configuration of frames in predicting the shot scale.
The performance of I3D-ResNet101 is similar to I3D-ResNet50 since deeper 3D networks needs more data to improve the performance.

From another perspective, 3D CNNs are much more computational expensive and need dense samples from videos, which causes the low speed for training and inference. We choose 2D TSN-ResNet50 as our backbone. The results prove that this 2D network can achieve better results than 3D networks with our careful designs.
Deep 2D network TSN~\cite{wang2016temporal} using image (TSN img) achieves $\sim30$\% raise on Acc$_{S}$ and Acc$_{M}$ than traditional methods, as it captures high-level semantic information such as the subject contours and the temporal relationship in a shot.

3) \textit{Deeper Backbones.}
To show that the improvement does not come from the increase 
of model parameters, we compare SGNet w/ Var (img + flow) (use ResNet50 backbone) with TSN-ResNet152 (img + flow).
SGNet w/ Var (img + flow) outperforms TSN-ResNet152 by
a margin of 2.62 on $Acc_S$ and 3.84 on $Acc_M$, with 15\%
fewer parameters and 19\% fewer GFLOPs.

4) \textit{2D Networks and Kinetics Pre-training.}
Our full model SGNet w/ Var (img + flow) which includes subject map guidance, motion information flow, and variance map, 
improves $3.49$ (relatively $4.12\%$) on $Acc_S$ and $11.40$ (relatively $16.11\%$)  on $Acc_M$ compared to TSN (img),
and $3.47$ (relatively $4.15\%$) on $Acc_S$ and $4.73$ (relatively $6.13\%$) on $Acc_M$ compared to TSN (img + flow).
The full model get further improvements by 0.2 on $Acc_S$ and 1.8 $Acc_M$ with Kinetics~\cite{carreira2017quo} pre-trained. This result shows that action recognition dataset can bring more help to shot movement predictions.\\

\noindent\textbf{Analysis of Our Framework.}
Based on TSN (img), SGNet (img) takes the advantage of subject map guidance and improves the scale and movement results by $3.13$ and $0.96$ respectively, which shows the usefulness of subject guidance especially for scale type prediction.
With the help of variance map,  SGNet w/ Var (img) raise the movement classification performance from $70.46$ to $80.57$ (relatively $14.35\%$).
Similarly, flow (SGNet img+ flow) helps the model to improve the movement results from $70.46$ to $80.65$ (relatively $14.46\%$). These results show that variance map and flow both capture the movement information and contribute to the great performance on movement type classification.
As for scale type classification, variance map and flow bring slight improvements ($0.2\sim0.3$), which shows that the movement information captured by variance map and flow provide a weak assistance to the scale type classification.
Finally, combining variance map and flow, SGNet w/ Var (img + flow) further gains improvement on scale ($87.57$) and on movement ($81.86$) classification and achieves the best performance among all (without Kinetics pre-trained).

\subsection{Ablation Studies}
We conduct ablation studies on the following designs to verify their effectiveness: 1) subject map guidance, 2) subject map generation, and 3) joint training.\\

\begin{table}[!t]
	\begin{center}
		\caption{
			Comparison of different subject or/and background map guidance.
		}
		\resizebox{0.61\columnwidth}{!}
		{
			\begin{tabular}{c|l|cc}
				\toprule
				\#&Settings& Acc$_{S}$ ($\uparrow$)& Acc$_{M}$ ($\uparrow$) \\ \midrule
				1&Base  (TSN w/ Var img+flow)    &  84.15    &    77.25    \\
				2&Subject only      &  79.97    &    74.65       \\
				3&Back only      &  79.60    &    75.80       \\ \midrule
				4&Base+Subj   &   \textbf{87.57}      &  80.86   \\
				5&Base+Back   &  87.10  &   \textbf{81.86}    \\
				6&Base+Subj+Back    &    87.31     &    81.54   \\
				\bottomrule
			\end{tabular}
		}
		\label{tab:subject}
	\end{center}
\end{table}

\noindent{\bf The Effects of Different Subject and Background Map Guidances.}
In the first block of Table~\ref{tab:subject}, we take TSN model using image and flow with variance map (TSN w/ Var img+flow) as our baseline to test the effects of different subject map guidances. It takes a single-branch ResNet50 as backbone and two models for image and flow respectively, and fuses their scores at the end.
We observe that using only subject or background information is inferior to the performance of using the whole image and flow, with $\sim5$/$\sim2$ drop on $Acc_{S}$/$Acc_{M}$.

Setting 4,5 in Table~\ref{tab:subject} are two branches setting with either subject or background guidance.
In these experiments, we take a two-branch ResNet50 as backbone for image and flow model, one branch for subject/background and the other one for the whole image/flow.
The output obtained from the first branch is concatenated with the output of second branch (+Subj and +Back) as guidance, and send to following networks.
Generally, subject guidance achieves $0.4\sim0.8$ performance gain on scale classification and background guidance outperforms subject guidance on movement prediction with $0.8\sim1.0$ better results.

Setting 6 (Base+Subj+Back) in Table~\ref{tab:subject} use a four-branch ResNet50. Two branches are for subject guidance, and the rest two are for background guidance.
With more information, the performance drops a little since the subject and background information might be mutually exclusive to each other. \\
%

\noindent{\bf The Influence of Different Subject Map Generations.}
As discussed above, a subject map generation module is needed to guide the network prediction. This module has many alternatives. Table~\ref{tab:atten} shows the comparisons between our approaches and self-attention SBS (Saliency-Based Sampling Layer)~\cite{recasens2018learning}, and fine-tuned/fixed models $R^3$Net-ResNet18/50~\cite{deng2018r3net}.
We take TSN model using image and flow with variance map (TSN w/ Var img+flow) as our baseline to test the influence of different subject map generations.

\begin{table}[!t]
	\begin{center}
		\caption{
			Comparison of different subject map generation modules.
		}
		\resizebox{0.71\columnwidth}{!}{%
		\begin{tabular}{c|l|cc}
			\toprule
			\#&Settings& Acc$_{S}$ ($\uparrow$)& Acc$_{M}$ ($\uparrow$) \\ \midrule
			1&Base  (TSN w/ Var img+flow)    &  84.15    &    77.25    \\
			2&SBS-ResNet50~\cite{recasens2018learning}      &  83.82    &    76.36    \\\midrule
			3&$R^3$Net-ResNet18-fixed~\cite{deng2018r3net}       &  84.55    &  78.14      \\
			4&$R^3$Net-ResNet50-fixed~\cite{deng2018r3net} &85.10 &79.56     \\ 
			5&$R^3$Net-ResNet18-finetuned~\cite{deng2018r3net}       &  86.15    &  81.24      \\
			6&$R^3$Net-ResNet50-finetuned~\cite{deng2018r3net} & \textbf{88.10}     & \textbf{82.58}         \\ \midrule \midrule
			7&Student generator w/ $\cL_{2}$ +  $\cL_{cross}$     & 85.34     & 79.11        \\
			8&Student generator w/ $\cL_{2}+\cL_{\mathrm{adv}}+ \cL_{cross}$ &   \textbf{87.08}        &  \textbf{81.13}    \\
			\bottomrule
		\end{tabular}
		}
		\label{tab:atten}
	\end{center}
\end{table}

Self-attention generation method SBS (Saliency-Based Sampling Layer)~\cite{recasens2018learning} does not bring improvement compared with the baseline. The reason might be that self-attention is hard to learn from these weak labels, \ie shot types.
The pre-trained fixed networks (settings 3,4) bring gains to the performance, and the performance increases as the network becomes deeper. Moreover, when we fine tune these networks on our tasks (settings 5,6), the performance improves further with $\sim2$ gains on both Acc$_S$ and Acc$_M$.

Our light-weight subject map generation module is driven by two losses besides the classification cross entropy loss. The performance of using $\cL_{2}$ loss (setting 7) is worse than fine-tuned $R^3$Net-ResNet50. With the help of both $\cL_{2}$ loss and adversarial loss $\cL_{adv}$ (setting 8), student generator is on par with $R^3$Net-ResNet50.
However, compared with $R^3$Net-ResNet50, our light-weight subject map student generator has 99.8\% fewer parameters and 89.4\% fewer GFLOPs (shown in Table~\ref{tab:attn-para}), which largely speeds up the training and inference processes.\\

\begin{table}[t]
	\begin{minipage}{0.49\textwidth}
		\centering 
		\caption{Parameters and computational complexity of different networks}
		\resizebox{0.95\textwidth}{!}
		{
		\begin{tabular}{l|cc}
			\toprule
			Network Architecture & Params(M) & GFLOPs \\ \midrule
			Student generator & \textbf{0.04} & \textbf{2.38} \\
			$R^3$Net-ResNet18~\cite{deng2018r3net} & 23.66 & 19.95 \\
			$R^3$Net-ResNet50~\cite{deng2018r3net} & 37.53 & 22.54 \\ \bottomrule
		\end{tabular}
		\label{tab:attn-para}
		}
	\medskip
	\end{minipage}
		\hfillx
	\begin{minipage}{0.49\textwidth}
		\centering
		\caption{
			Comparison of the performance of joint training sharing different modules.}
		\resizebox{\textwidth}{!}
		{
		\begin{tabular}{l|ccl}
	\toprule
	Settings & Acc$_{S}$ ($\uparrow$)& Acc$_{M}$ ($\uparrow$) \\ \midrule
	Separate       &  87.57    &  81.86      \\
	Joint-training (Share SMG)  & \textbf{88.12} & \textbf{82.19}\\
	Joint-training (Share till res$_1$)    &  87.24       &   81.10               \\
	Joint-training (Share till res$_4$)    &  86.17       &    80.29	              \\
	\bottomrule
\end{tabular}
		}
	\label{tab:joint}
	\end{minipage}
\end{table}

\noindent{\bf Two-task Joint Training.}
To investigate the relationship between the scale and movement classification,
we conduct the joint training experiments on these two tasks,
as shown in Table~\ref{tab:joint}. We take our full model SGNet w/ Var (img~+~flow) as the baseline.
We testify the coupling of
scale and movement type by sharing the same modules from bottom to top gradually.
As lined out in the second row, sharing the subject map generation (SMG) module is helpful to the performance, where $Acc_S$ and $Acc_M$ raises $0.55$ and $0.33$ respectively. 
However, when we further share these two tasks classification till ResNet50's {res$_1$} and {res$_4$} modules, we observe that joint training is harmful to the performance when they share more branches.
These prove that both scale and movement benefit from the subject guidance. The spatial layout learnt from scale and the camera motion learnt from the movement contribute complementaily to the subject map generation. While the subject guidance is shared by both tasks, the distinct goals of the two tasks still require task-specific designs in the later part to learn better representations.

\subsection{Qualitative Results}
In this section, we show the qualitative results of subject map generation and the variance map computation.\\

\begin{figure}[!t]
	\begin{center}
		\includegraphics[width=0.90\linewidth]{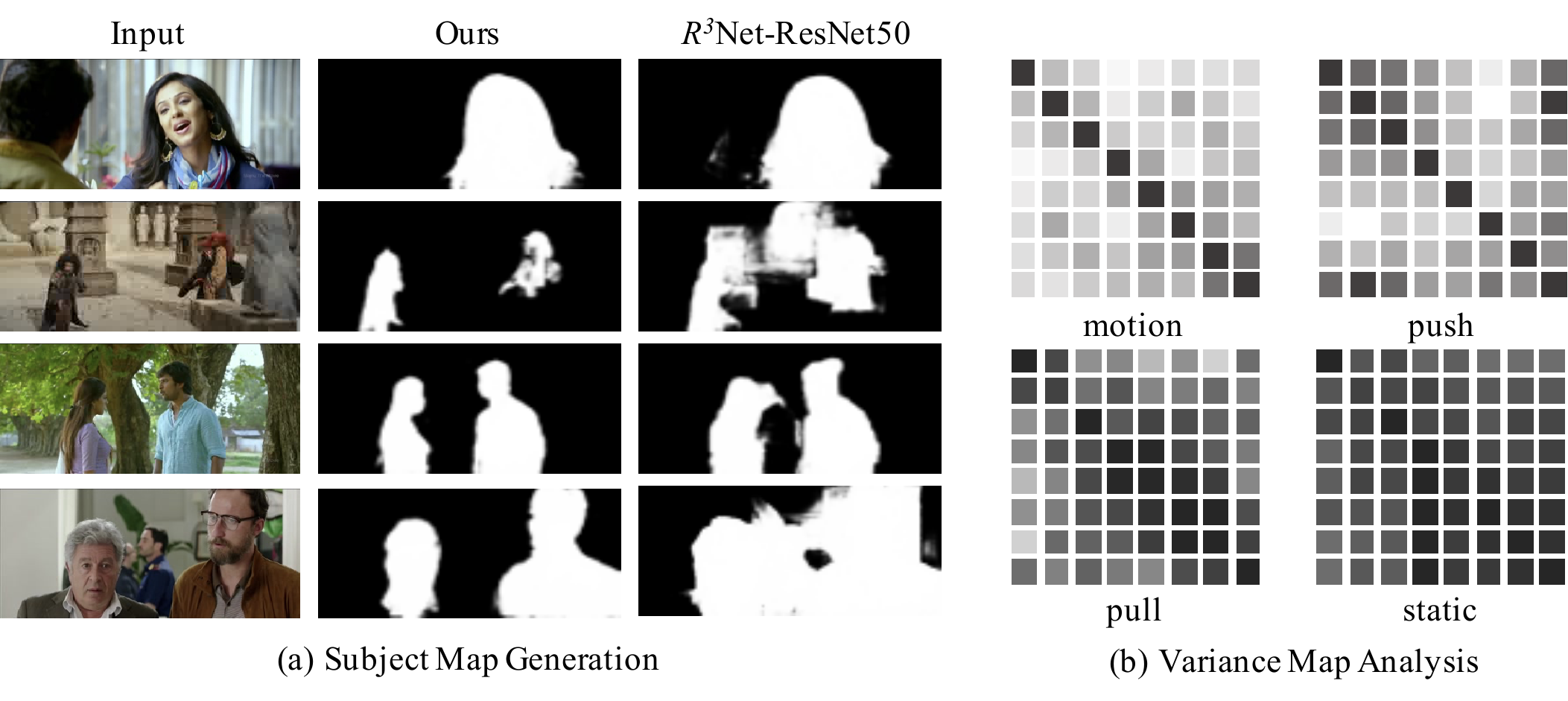}
	\end{center}
	\caption{
		(a) Comparison of our generated subject map and $R^3$Net-ResNet50-fixed generated map. (b) Variance map visualization of different movement types using gray-scale colors to indicate the similarity. The lighter the color, the lower the similarity score
	}
	\label{fig:var}
\end{figure}

\noindent{\bf Subject Map.}
Fig.~\ref{fig:var}(a) compares our generated subject map with those generated by $R^3$Net-ResNet50 in fixed setting.
Our generated subject map achieves much better generation result that are consistent with our human judgment.
The first row in the figure is an over-the-shoulder static close-up shot, where both methods successfully predict the subject woman rather than the man with back head.
But our method outputs much less noise.
The second row is a full shot. Our method successfully detects both two people and does not include the background stone into the subject map.
In the third and fourth row cases, $R^3$Net-ResNet50-fixed outputs blurred area around the contours of two people while our method obtains a sharp shape of the subject.\\

\noindent{\bf Variance Map.}
The variance map is important for predicting shot movement.
We divide a shot video into 8 clips, plot the variance map for each movement type in the test set and average these variance maps in Fig.~\ref{fig:var}(b).
The variance map is of size $8\times8$, and these gray scale blocks show the similarity among clips in the variance map.
As noted from the plot, the variance map of the static shot is nearly an all-one matrix, meaning that there is no significant change between the eight clips.
The near identity matrix shape of motion shots reveal that it has the least similarities between consecutive clips.\\

\section{Application}
\label{sec:application}
 
Shot type analysis has a wide range of potential applications. In this section, we illustrate one such application of realizing automatic video editing with the help of shot type classification.\\

\noindent \textbf{Automatic Video Editing: Shot Type Changing.}
Video editors usually try out different shot types to convey emotions and stories, which consumes a lot of time and resources.
In many cases, the change of shot type changes the semantic of a movie and effects audience's emotions, revealing the intent of the directors. 

\begin{figure}[!t]
	\hspace{4pt} \includegraphics[width=0.99\linewidth]{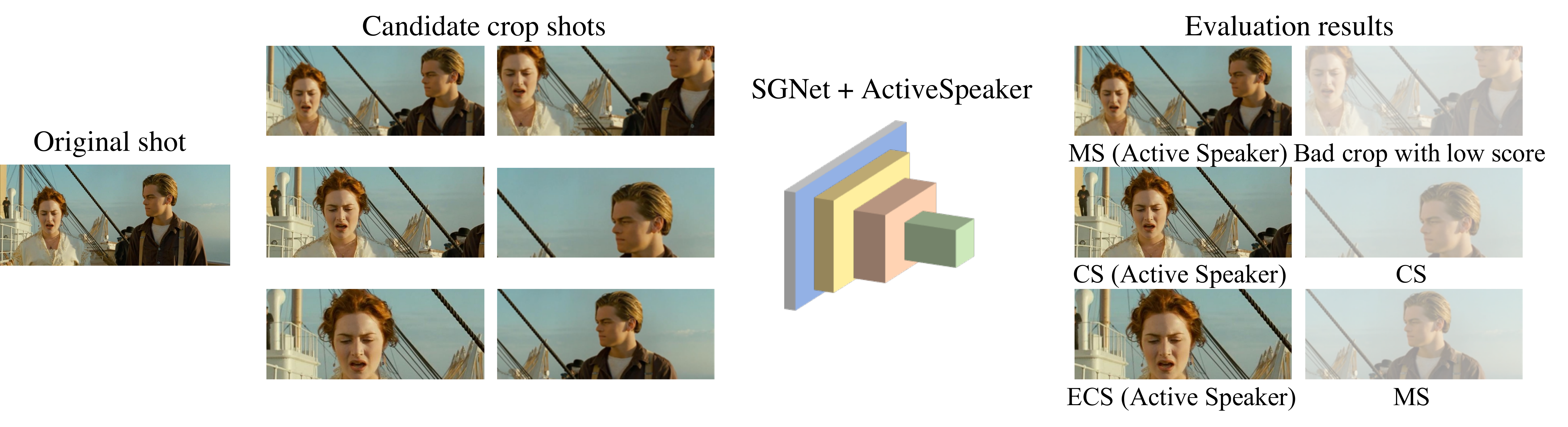}
	\caption{\small
		A sample shot from a dialogue scene in \emph{Titanic}, showing how we use our proposed shot type classification framework to aid the shot type changing
	}
	\label{fig:crop}
\end{figure}

\begin{figure}[!t]
	\begin{center}
		\includegraphics[width=0.99\linewidth]{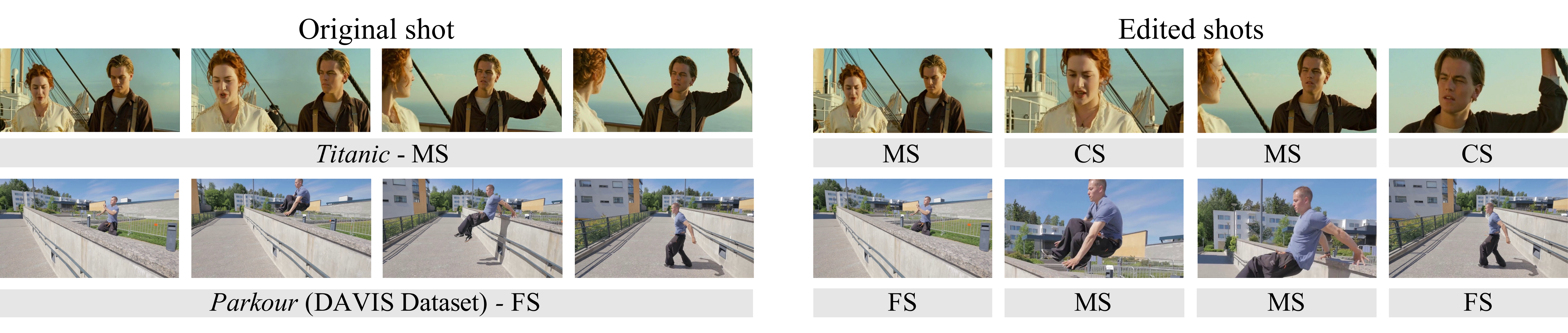}
	\end{center}
	\caption{
		Editing results on a medium shot clip from \emph{Titanic} to emphasize the speaker and on a full shot clip \emph{Parkour} from the DAVIS dataset~\cite{Caelles_arXiv_2019} to emphasize the action}
	\label{fig:davis}
\end{figure}

The model we propose in this paper could classify the shot type of any given video shot. Fig.~\ref{fig:crop} shows a shot clip\footnote[3]{
~\cite{sidiropoulos2011temporal} is adopted here to cut shots from the film.} from the famous film \emph{Titanic}, demonstrating how our model can be applied to changing shot scales to achieve a desired artistic expression.
The original shot is a \emph{medium shot}. Suppose we want to emphasize the role of speaker in this \emph{dialogue scene}, we may want to use some \emph{close-up shot} to emphasize the speaker.
Firstly, we propose many cropping regions randomly depend on the position of the speaker and generate the corresponding candidate shots.
Secondly, we use our shot classification model to classify these candidate shots and assign them with confidence scores. 
Note that the original shot is a single shot. We divide it into four shots depending on the active-speaker~\cite{roth2019ava}. In Shot 1, Rose and Jack walk on the deck of the ship; Shot 2, Rose talks; Shot 3, they stop and Rose looks at Jack; Shot 4, Jack talks, as illustrated in Fig.~\ref{fig:crop}. 
We change the style of the original shot by selecting parts from the divided four shots and replace them with the candidates with high scores and the desired scale types, as shown in Fig.~\ref{fig:davis}.
After these changes, the emotion of this clip turns to be more intense and the speaking cast is being emphasized after changing from a middle shot to a close shot. One more result on DAVIS dataset~\cite{Caelles_arXiv_2019} is also shown in Fig.~\ref{fig:davis}. These results demonstrate the importance of shot type in videos, especially for their emotion and aesthetic analysis.\footnote[4]{More results and their corresponding videos are shown in the supplementary videos.} 


\section{Conclusion}
\label{sec:conclusion}
In this work, we construct a large-scale dataset \emph{MovieShots} for shot analysis, which containing $46K$ shots from $7K$ movie trailers with professionally annotated scale and movement attributes.
We propose a Subject Guidance Network (SGNet) to capture the contextual information and the spatial and temporal configuration of a shot for our shot type classification task.
Experiments show that this network is very effective and achieves better results than existing methods. 
All the studies in this paper together show that shot type analysis is a promising direction for edited video analysis which deserves further research efforts.\\

\noindent\textbf{Acknowledgement}: This work is partially supported by the SenseTime Collaborative Grant on Large-scale Multi-modality Analysis (CUHK Agreement No. TS1610626 \& No. TS1712093), the General Research Fund (GRF) of Hong Kong (No. 14203518 \& No. 14205719), and Innovation and Technology Support Program (ITSP) Tier 2, ITS/431/18F.

\clearpage
%
%
\bibliographystyle{splncs04}
\bibliography{main}
\end{document}